\documentclass[runningheads]{llncs}
\usepackage{graphicx}
\usepackage[table]{xcolor}
\usepackage{comment}
\usepackage{amsmath,amssymb} % define this before the line numbering.
\usepackage{subfig}
\usepackage{microtype}
\usepackage{booktabs}
\usepackage{nicefrac}
\usepackage{multirow}
\usepackage{tabularx}
\usepackage{orcidlink}
\usepackage{hyphenat}
\usepackage{footnote}
\makesavenoteenv{tabular}
\makesavenoteenv{table}

\usepackage[accsupp]{axessibility}  % Improves PDF readability for those with disabilities.

\DeclareMathOperator*{\argmin}{arg\,min}

\definecolor{mathieu_color}{rgb}{.6,.4,.05}
\definecolor{henrique_color}{rgb}{0,0.35,0}
\definecolor{JF_color}{rgb}{0,0,0.35}

\definecolor{best}{RGB}{255, 220, 200}
\definecolor{second}{RGB}{255, 255, 200}
\begin{document}
% \linenumbers
\pagestyle{headings}
\mainmatter
\def\ECCVSubNumber{2772}  % Insert your submission number here

\title{Editable indoor lighting estimation} % Replace with your title

\begin{comment}
\titlerunning{ECCV-22 submission ID \ECCVSubNumber} 
\authorrunning{ECCV-22 submission ID \ECCVSubNumber} 
\author{Anonymous ECCV submission}
\institute{Paper ID \ECCVSubNumber}
\end{comment}
%******************

%\begin{comment}
\titlerunning{Editable indoor lighting estimation}
\author{Henrique Weber\inst{1}\orcidlink{0000-0001-7103-7220} \and
Mathieu Garon\inst{2}\orcidlink{0000-0003-1811-4156} \and
Jean-Fran\c{c}ois Lalonde\inst{1}\orcidlink{0000-0002-6583-2364}}
% \author{Henrique Weber\inst{1} \and
% Mathieu Garon\inst{2} \and
% Jean-Fran\c{c}ois Lalonde\inst{1}}
%
\authorrunning{H. Weber et al.}
\institute{Université Laval, Québec, Canada \and
Depix, Montréal, Canada \\
\small{\texttt{\url{https://lvsn.github.io/EditableIndoorLight/}}}}

%\end{comment}
%******************

\maketitle

\begin{abstract}
We present a method for estimating lighting from a single perspective image of an indoor scene. Previous methods for predicting indoor illumination usually focus on either simple, parametric lighting that lack realism, or on richer representations that are difficult or even impossible to understand or modify after prediction. We propose a pipeline that estimates a parametric light that is easy to edit and allows renderings with strong shadows, alongside with a non-parametric texture with high-frequency information necessary for realistic rendering of specular objects. Once estimated, the predictions obtained with our model are interpretable and can easily be modified by an artist/user with a few mouse clicks. Quantitative and qualitative results show that our approach makes indoor lighting estimation easier to handle by a casual user, while still producing competitive results.

\end{abstract}

%!TEX root = eccv2022submission.tex
\section{Introduction}

Mixing virtual content realistically with real imagery is required in an increasing range of applications, from special effects to image editing and augmented reality (AR). This has created the need for capturing the lighting conditions of a scene with ever increasing accuracy and flexibility. In his seminal work, Debevec~\cite{debevec-sig-98} suggested to capture the lighting conditions with a high dynamic range light probe. While it has been improved over the years, this technique, dubbed \emph{image-based lighting}, is still at the heart of lighting capture for high end special effects in movies nowadays\footnote{See {\scriptsize \url{https://www.fxguide.com/fxfeatured/the-definitive-weta-digital-guide-to-ibl/}}.}. Since the democratization of virtual object insertion for consumer image editing and AR, capturing light conditions with light probes restricts non professional users to have access to the scene and to use specialized equipment. To circumvent those limitations, approaches for automatically estimating the lighting conditions directly from images have been proposed. 

In this line of work, the trend has been to estimate more and more \emph{complex} lighting representations. This is exemplified by works such as Lighthouse~\cite{srinivasan2020lighthouse}, which propose to learn a multi-scale volumetric representation from an input stereo pair. Similarly, Li et al.~\cite{li2020inverse} learn a dense 2D grid of spherical gaussians over the image plane. Wang et al.~\cite{wang2021learning} propose to learn a 3D volume of similar spherical gaussians. While these lighting representations have been shown to yield realistic and spatially-varying relighting results, they have the unfortunate downside of being hard to understand: they do not lend themselves to being easily editable by a user. This quickly becomes a source of limitation when erroneous automatic results need to be corrected for improved accuracy or when creative freedom is required.

\begin{figure}[t]
\centering
\includegraphics[width=\textwidth]{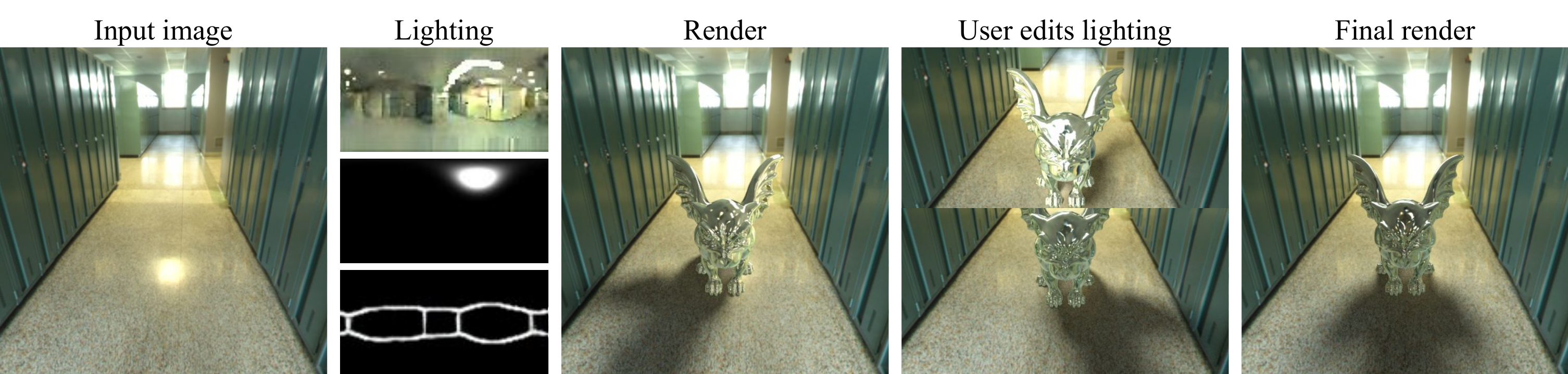}
\caption{Our method produces an estimation of the indoor lighting from a single perspective image. Our lighting representation is composed of a 3D parametric light source, a texture map and a coarse 3D layout of the scene. With this information, it is possible to realistically insert 3D objects (like the golden armadillo and sphere) into the scene. Because our lighting representation is interpretable and intuitive, the user can experiment with possibilities by modifying, say, the position of the light source in order to achieve the desired look.}
\label{fig:teaser}
\end{figure}

In this work, we depart from this trend and propose a \emph{simple, interpretable, and editable} lighting representation (fig.~\ref{fig:teaser}). 
But what does it mean for a lighting representation to be editable? We argue that an editable lighting representation must: 1) \textit{disentangle} various components of illumination; 2) allow an \textit{intuitive control} over those components; and, of course, 3) enable \textit{realistic relighting results}.
Existing lighting representations in the literature do not possess all three properties. \emph{Environment maps}~\cite{gardner2017learning,song2019neural,legendre2019deeplight} can be rotated but they compound light sources and environment textures together such that one cannot, say, easily increase the intensity of the light source without affecting everything else. 
Rotating the environment map inevitably rotates the entire scene, turning walls into ceilings, etc., when changing the elevation.
\emph{Dense and/or volumetric} representations~\cite{garon2019fast,li2020inverse,srinivasan2020lighthouse,wang2021learning} are composed of 2D (or 3D) grids containing hundreds of parameters, which would have to be modified in a consistent way to achieve the desired result, an unachievable task for most. \emph{Parametric} representations~\cite{Gardner_2019_ICCV} model individual light sources with a few intuitive parameters, which can be modified independently of the others, but cannot generate realistic reflections.

Our proposed representation is the first to offer all three desired properties and is composed of two parts: 1) a parametric light source for modeling shading in high dynamic range; and 2) a non-parametric texture to generate realistic reflections off of shiny objects. 
Our representation builds on the hypothesis (which we validate) that most indoor scenes can accurately be modeled by a \emph{single}, dominant directional light source. We model this in high dynamic range with a parametric representation~\cite{Gardner_2019_ICCV} that explicitly models the light source intensity, size, and 3D position. This representation is intuitive and can easily be edited by a user simply by moving the light source around in 3D. 

This light source is complemented with a spatially-varying environment map texture, mapped onto a coarse 3D representation of the indoor scene. For this, we rely on a layout estimation network, which estimates a cuboid-like model of the scene from the input image. In addition, we also use a texture estimation network, whose output is conditioned on a combination of the input image, the scene layout and the parametric lighting representation. By explicitly tying the appearance of the environment texture with the position of the parametric light source, modifying the light source parameters (e.g. moving around the light) will automatically adjust the environment in a realistic fashion.

While our representation is significantly simplified, we find that it offers several advantages over the previous approaches. First, it renders both realistic shading (due to the high dynamic range of the estimated parametric light) and reflections (due to the estimated environment map texture). Second, it can efficiently be trained on real images, thereby alleviating any domain gap that typically arise when approaches need synthetic imagery for training~\cite{srinivasan2020lighthouse,li2020inverse,wang2021learning}. Third---and perhaps most importantly---it is \emph{interpretable and editable}. Since all automatic approaches are bound to make mistakes, it is of paramount importance in many scenarios that their output be adjustable by a user. By modifying the light parameters and/or the scene layout using simple user interfaces, our approach bridges the gap between realism and editability for lighting estimation. 

%!TEX root = eccv2022submission.tex
\section{Related work}

For succinctness, we focus on single-image indoor lighting estimation methods in the section below, and refer the reader to the recent survey on deep models for lighting estimation for a broader overview~\cite{einabadi2021deep}. 

\subsubsection{Lighting estimation}
Gardner et al.~\cite{gardner2017learning} proposed the first deep learning-based lighting estimation method for indoor scenes, and predicted an HDR environment map (equirectangular image) from a single image. This representation was also used in \cite{legendre2019deeplight} for both indoors and outdoors, in \cite{song2019neural} to take into account the object insertion position, in \cite{somanath2021hdr} which presented a real-time on-device approach, in \cite{sengupta2019neural} for scene decomposition, and in \cite{cheng2018learning} which exploited the front and back cameras in current mobile devices. Finally, \cite{weber2018learning} propose to learn the space of indoor lighting using environment maps on single objects.

Other works explored alternative representations, such as spherical harmonics~\cite{garon2019fast,mandl2017learning,zhao2020pointar} that are useful for real-time rendering but are typically unsuitable for modeling high-frequency lighting (such as bright light sources) and are not ideal for non diffuse object rendering. \cite{Gardner_2019_ICCV} proposed to estimate a set of 3 parametric lights, which can easily be edited. However, that representation cannot generate realistic reflections. EMlight~\cite{zhan2021emlight} propose a more expressive model by predicting gaussians on a spherical model. Similar to us, GMlight~\cite{zhan2021gmlight} back-projects the spherical gaussians to an estimated 3D model of the scene. This is further extended in \cite{bai2022deep} by the use of graph neural networks, and in \cite{zhan2021sparse} through the use of spherical wavelets dubbed ``needlets''. %In contrast, our method combines a parametric light with an environment map texture that is back-projected onto a coarse 3D model of the scene. Critically, the generated texture is conditioned on the input light and 3D model, which allows us to intuitively control the estimated lighting and geometry parameters while simultaneously generating realistic environments. 

Recently, methods have attempted to learn volumetric lighting representations from images. Of note, Lighthouse~\cite{srinivasan2020lighthouse} learns multi-scale volumetric lighting from a stereo pair, \cite{li2020inverse} predicts a dense 2D grid of spherical gaussians which is further extended into a 3D volumetric representation by Wang et al.~\cite{wang2021learning}. While these yield convincing spatially-varying results, these representations cannot easily be interacted by a user. 

\subsubsection{Scene decomposition}
Holistic scene decomposition~\cite{barron2014shape} is deeply tied to lighting estimation as both are required to invert the image formation process. Li et al.~\cite{li2020inverse} proposes to extract the scene geometry and the lighting simultaneously. Similarly, \cite{eigen2015predicting} extract only the geometry of the scene by estimating the normal and depth of the scene. These geometric representations are however non-parametric and thus difficult to edit or comprehend.
\cite{lee2017roomnet} proposes a simplified parametric model where a room layout is recovered in the camera field of view. Similarly, \cite{zou2018layoutnet} presents a method to estimate the layout given a panoramic image of an indoor scene. We use the method of \cite{lee2017roomnet} to estimate a panoramic layout given a perspective image, thus providing a simple cuboid representation that allows for spatially varying textured lighting representation.
%!TEX root = eccv2022submission.tex
\section{Editable indoor lighting representation}
\label{sec:lighting-representation}

We begin by presenting our hybrid parametric/non-parametric lighting representation which aims at bridging the gap between realism and editability. We also show how that representation can be fitted to high dynamic range panoramas to obtain a training dataset, and conclude by presenting how it can be used for virtual object relighting.

\subsection{Lighting representation}

Our proposed light representation, shown in fig.~\ref{fig:rendering_pipeline}, is composed of two main components: an HDR parametric light source $\mathbf{p}$; and an LDR textured cuboid $\mathcal{C}$. 

\subsubsection{Light source}
As in \cite{Gardner_2019_ICCV}, the light source parameters $\mathbf{p}$ are defined as
\begin{equation}
    \mathbf{p} = \{\mathbf{l}, d, s, \mathbf{c}, \mathbf{a}\}\,,
\end{equation}
where $\mathbf{l} \in \mathbb{R}^3$ is a unit vector specifying the light direction in XYZ coordinates, $d$ is the distance in meters, $s$ the radius (in meters), $\mathbf{c}, \mathbf{a} \in \mathbb{R}^3$ are the light source and ambient colors in RGB, respectively. Here, $\mathbf{l}$, $d$ and $s$ are defined with respect to the camera. In contrast with \cite{Gardner_2019_ICCV}, we use a single light source. 

\subsubsection{Textured cuboid}

The cuboid $\mathcal{C} = \{\mathbf{T}, \mathbf{L}\}$ is represented by a texture $\mathbf{T} \in \mathbb{R}^{2H \times H \times 3}$, which is an RGB spherical image of resolution $2H \times H$ stored in equirectangular (latitude-longitude) format, and a scene layout $\mathbf{L} \in \mathbb{R}^{2H \times H}$. The layout is a binary image of the same resolution, also in equirectangular format, indicating the intersections of the main planar surfaces in the room (walls, floor, ceiling) as an edge map~\cite{fernandez2020corners}.

\begin{figure}[t]
\centering
\includegraphics[width=\textwidth]{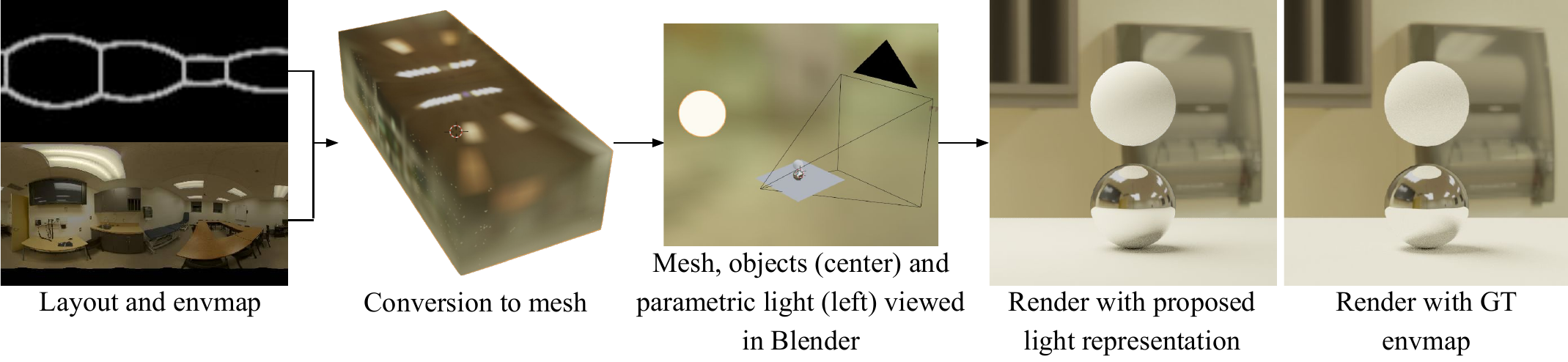}
\caption{To render a virtual object with our proposed lighting representation, the texture is first warped according to the layout (1st column), producing a textured mesh (2nd). This mesh is combined with an emitting sphere representing the parametric light (3rd) for rendering. The resulting rendering (4th) closely matches the ground truth rendering obtained with the HDR environment map (last). }
\label{fig:rendering_pipeline}
\end{figure}

\subsection{Ground truth dataset}
\label{sec:ground-truth-dataset}

The ground truth is derived from the Laval Indoor HDR Dataset~\cite{gardner2017learning}, which contains 2,100 HDR panoramas (with approximate depth labels from \cite{Gardner_2019_ICCV}). We extract $\mathbf{p}$ and $\mathcal{C}$ from each panorama using the following procedure. First, the HDR panorama is clipped to LDR (we re-expose such that the 90th-percentile is 0.8 then clip to [0, 1]) and directly used as the texture $\mathbf{T}$. Then the intersection between the main surfaces are manually labelled to define the layout $\mathbf{L}$. Lastly, we extract a dominant parametric light source from the HDR panorama. In order to determine the main light source, the $N=5$ brightest individual light sources are first detected using the region-growing procedure in \cite{Gardner_2019_ICCV}. A test scene (9 diffuse spheres arranged in a $3 \times 3$ grid on a diffuse ground plane, seen from top as in fig.~\ref{subfig:strlight-images}) is rendered with each light source independently by masking out all other pixels---the brightest render determines the strongest light source. 

% The region is kept for each light sources, masking out other values. Given each light regions, we determine the light contribution of each light by rendering a test scene (9 diffuse spheres arranged in a $3 \times 3$ grid on a diffuse ground plane, seen from top as in fig.~\ref{subfig:strlight-images}). The region transmitting the highest energy to the scene is kept. 

An initial estimate of the light parameters $\mathbf{p}$ are obtained by the following. The distance $d$ is approximated by using the average depth of the region, direction $\mathbf{l}$ as the region centroid, the angular size from the major and minor axes of an ellipse fitted to the same region. Finally, the light color $\mathbf{c}$ and ambient term $\mathbf{a}$ are initialized with a least-squares fit to a rendering of the test scene using the HDR panorama. From the initial parameters,  $\mathbf{p}$ is further refined:
\begin{equation}
\mathbf{p}^* = \argmin_{\mathbf{p}} ||\mathcal{R}(\mathbf{p}) - \mathcal{R}(\tilde{\mathbf{P}})||_2 \,.
\end{equation}
$\mathcal{R}(x)$ is a differentiable rendering operator (implemented with Redner~\cite{li2018differentiable}) that renders a test scene using $\mathbf{p}$. The optimization is performed using gradient descent with Adam~\cite{kingma2014adam}. Finally, the texture map $\mathbf{T}$ is rescaled with the estimated ambient term $\mathbf{a}^*$ to ensure that the texture yields the same average RGB color.

\subsection{Virtual object rendering}

To render a virtual object using our lighting representation, we employ the Cycles rendering engine\footnote{Available within Blender at \url{https://www.blender.org}.}. A scene, as shown in fig.~\ref{fig:rendering_pipeline}, is composed of a 3D emissive sphere for the parametric light $\mathbf{p}$ and the textured cuboid mesh $\mathcal{C}$. The cuboid mesh is derived by detecting the cuboid corners from the layout using high pass filters. We use the following geometric constraints to simplify the back-projection of the scene corners to 3D. First, the shape is limited to a cuboid, meaning that opposing faces are parallel. Second, the panorama layouts were trained using a camera elevation of $0^\circ$ (pointing at the horizon) and height of 1.6 meter above the ground. Using these constraints, the bottom corners can easily be projected on the ground plane, and the top corners can be used to compute the ceiling height (averaged from the 4 corners). A texture map can then be computed using every planar surfaces of the cuboid. Finally, the parametric light and the texture are rendered in two rendering passes. After rendering, the relit virtual object can be composited into the image using differential rendering~\cite{debevec-sig-98}. 

%!TEX root = eccv2022submission.tex
\section{Approach}
\label{sec:approach}

\begin{figure}[t]
\centering
\includegraphics[width=\textwidth]{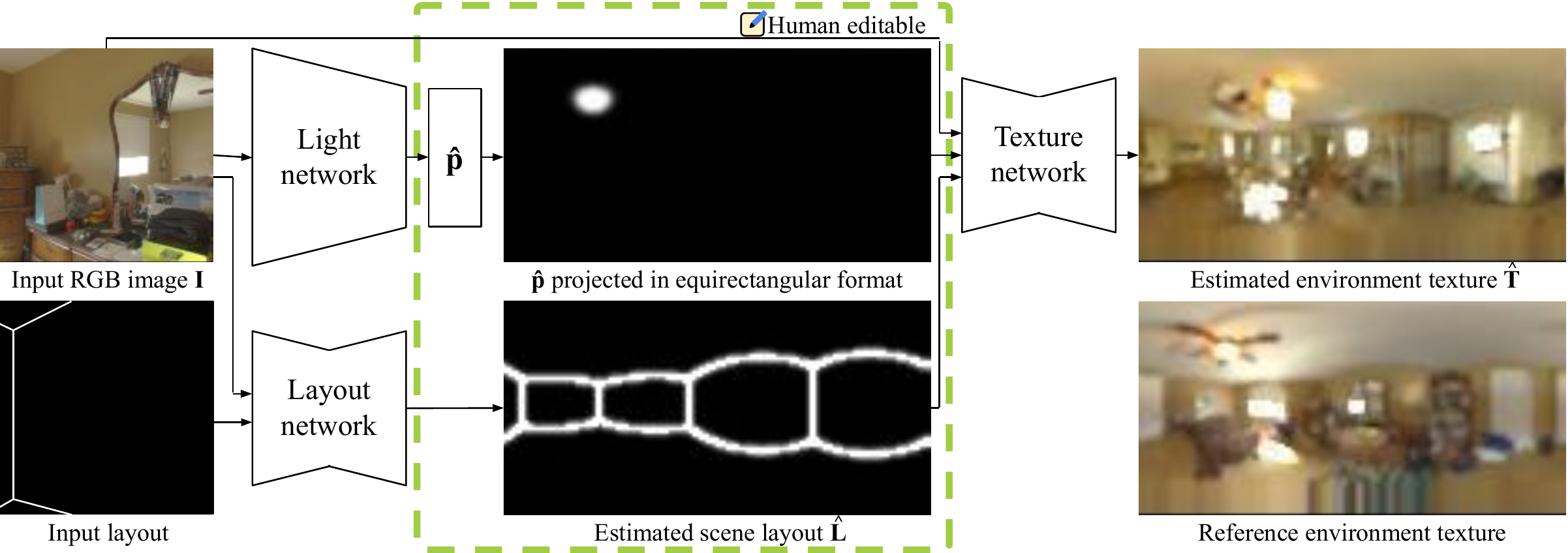}
\caption{Our method takes as input a perspective, RGB image and its scene layout representation, passes the RGB to a CNN to predict a parametric light, and passes the partial layout to another CNN to predict the full panorama layout. The parametric light is converted to a binary mask panorama, which is then sent together with the full layout prediction and the input RGB image to a third network which outputs an LDR texture with the light at the desired location. }
\label{fig:pipeline}
\end{figure}

Our approach, illustrated in fig.~\ref{fig:pipeline}, is composed of three main networks: light, layout, and texture which are combined together to estimate our light representation (c.f., sec.~\ref{sec:lighting-representation}) from an image. We assume that the layout of the input image is available, in practice this is obtained with an off-the-shelf solution \cite{yang2022learning}.

\subsubsection{Light network}

A ``light'' network is trained to learn the mapping from input image $\mathbf{I} \in \mathbb{R}^{128 \times 128 \times 3}$ to estimated lighting parameters $\mathbf{p}$ (sec.~\ref{sec:lighting-representation}) using a similar approach to \cite{Gardner_2019_ICCV}. Specifically, the light network is composed of a headless DenseNet-121 encoder~\cite{huang2017densely} to produce a 2048-dimensional latent vector, followed by a fully-connected layer (512 units), and ultimately with an output layer producing the light source parameters $\mathbf{p}$. 

The light network is trained on light parameters fitted on panoramas from the Laval Indoor HDR Dataset~\cite{gardner2017learning} using the procedure described in sec.~\ref{sec:ground-truth-dataset}. To generate the input image from the panorama, we follow \cite{gardner2017learning} and extract rectified crops from the HDR panoramas. The resulting images are converted to LDR by re-exposing to make the median intensity equal to 0.45, clipping to 1, and applying a $\gamma=\nicefrac{1}{2.4}$ tonemapping. The same exposure factor is subsequently applied to the color $\mathbf{c}$ and ambient $a$ light parameters to ensure consistency. Note that the training process is significantly simplified compared to \cite{gardner2017learning} as the network predicts only a single set of parameters.

We employ individual loss functions on each of the parameters independently: L2 for direction $\mathbf{l}$, depth $d$, size $s$, and ambient color $a$, and L1 for light color $\mathbf{c}$. In addition, we also employ an angular loss for both the ambient and light colors $a$ and $\mathbf{c}$ to enforce color consistency. The weights for each term were obtained through a Bayesian optimization on the validation set (see supp. mat.). 

\subsubsection{Layout network}

The mapping from the input RGB image $\mathbf{I}$ and its layout (obtained with \cite{yang2022learning}) to the estimated scene layout $\hat{\mathbf{L}}$ (sec.~\ref{sec:lighting-representation}) is learned by the ``layout'' network whose architecture is that of pix2pixHD~\cite{wang2018high}. Both inputs are concatenated channel-wise. The layout network is trained on both the Laval and the Zillow Indoor Dataset~\cite{ZInD}, which contains 67,448 LDR indoor panoramas of 1575 unfurnished residences along with their scene layouts. To train the network, a combination of GAN, feature matching and perceptual losses are employed~\cite{wang2018high}. The same default weights as in \cite{wang2018high} are used in training. 

\subsubsection{Texture network}

Finally, the estimated environment texture $\hat{\mathbf{T}}$ is predicted by a ``texture'' network whose architecture is also that of pix2pixHD~\cite{wang2018high}. It accepts as input a channel-wise concatenation of three images: the input RGB image $\mathbf{I}$, the estimated light parameters $\hat{\mathbf{p}}$ projected in an equirectangular format, and the estimated scene layout $\hat{\mathbf{L}}$. The equirectangular images are vertically concatenated to the input image. Note that the $\hat{\mathbf{p}}$ projection is performed using a subset of all parameters (direction $\mathbf{l}$ and size $s$ only). 

The texture network is also trained on both Laval and Zillow datasets. To obtain the required light source position from the Zillow dataset, we detect the largest connected component whose intensity is above the 98th percentile over the upper half of the panorama. To convert the Laval HDR panoramas to LDR, first a scale factor is found such as the crop taken from that panorama has its 90th percentile mapped to 0.8. This scale factor is then applied to the panorama such as its scale matches the one of the crop. The texture network is trained with the same combination of losses as the layout network.

%!TEX root = eccv2022submission.tex
%!TEX root = eccv2022submission.tex
\begin{figure}[ht!]
\centering
\scriptsize
\subfloat[\label{subfig:strlight-fig}]{
\begin{tabular}{c}
\includegraphics[width=0.48\linewidth]{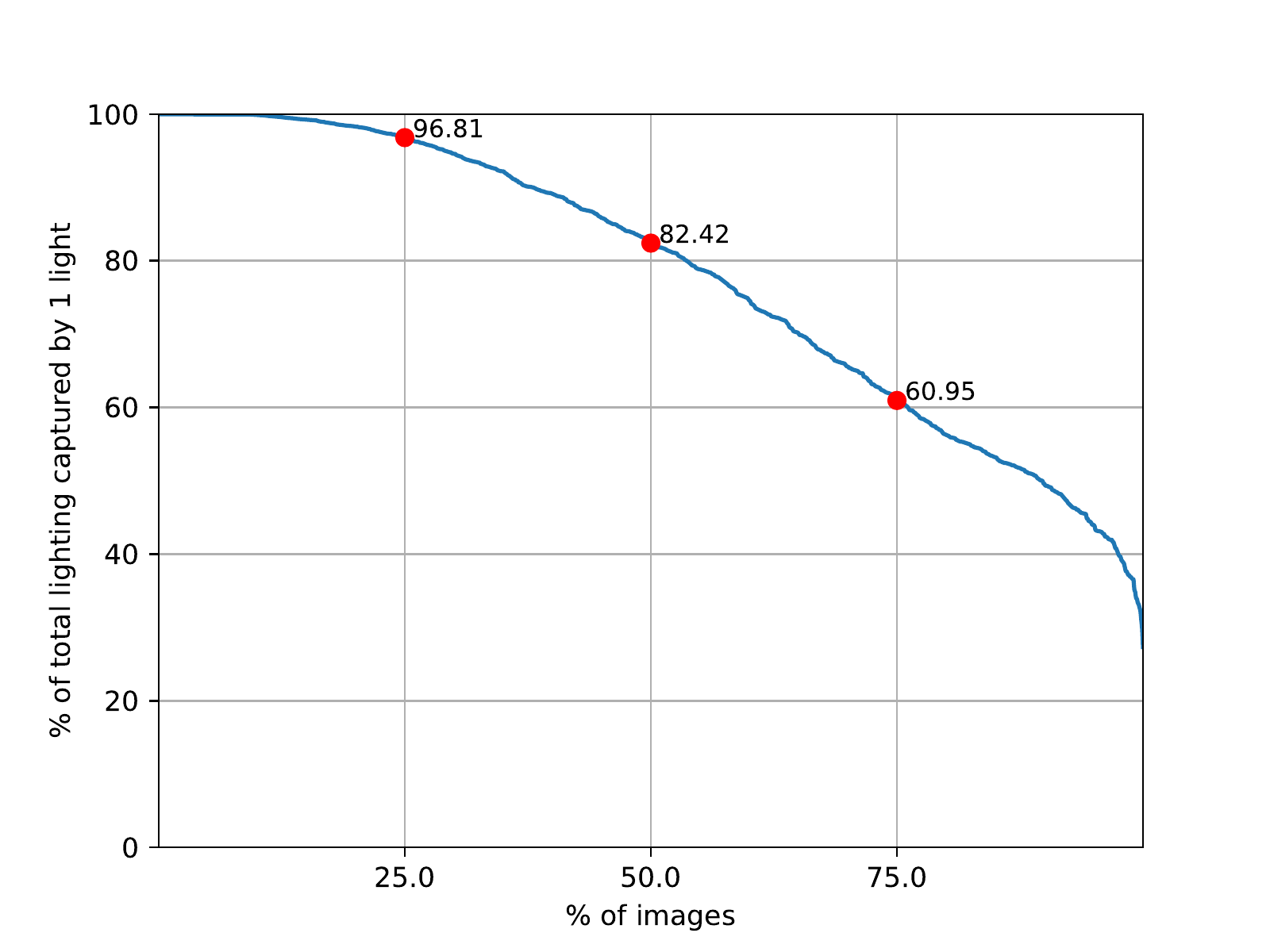}
\end{tabular}
}
\subfloat[\label{subfig:strlight-images}]{
\setlength{\tabcolsep}{1pt}
\begin{tabular}{cccc}
& 
% 1st prct. & 
25th prct. & 
50th prct. & 
75th prct. \\ 
& 
\includegraphics[width=0.13\linewidth]{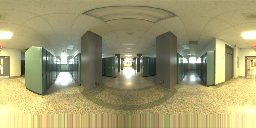} &
\includegraphics[width=0.13\linewidth]{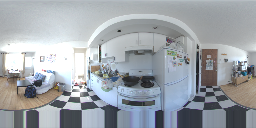} &
\includegraphics[width=0.13\linewidth]{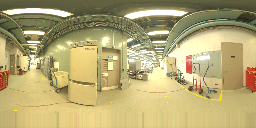} \\
\rotatebox{90}{\hspace{1.8em}GT} & 
\includegraphics[width=0.13\linewidth]{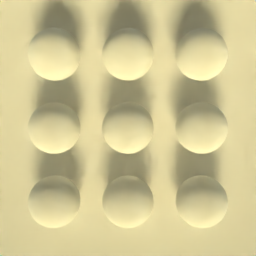} &
\includegraphics[width=0.13\linewidth]{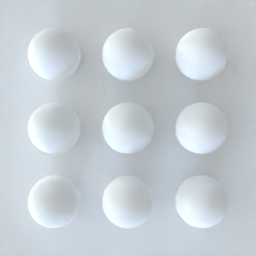} &
\includegraphics[width=0.13\linewidth]{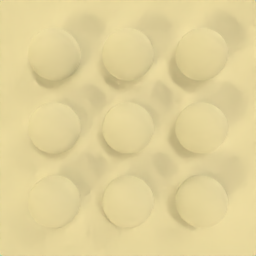} \\
\rotatebox{90}{\hspace{1.5em}Ours} & 
\includegraphics[width=0.13\linewidth]{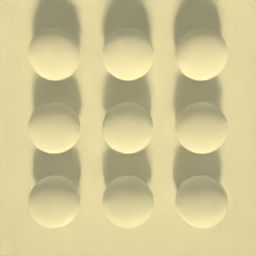} &
\includegraphics[width=0.13\linewidth]{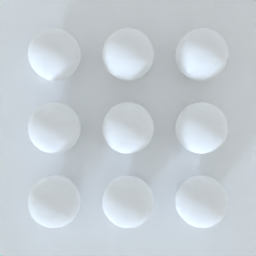} &
\includegraphics[width=0.13\linewidth]{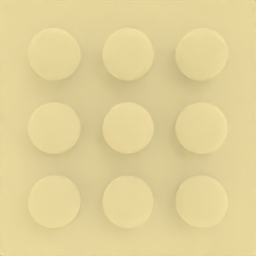} 
\end{tabular}
}
\caption[]{Validation of our 1-light approximation. \subref{subfig:strlight-fig} Cumulative distribution of the contribution of the single strongest light with respect to the entire lighting environment of the scene. \subref{subfig:strlight-images} Example images for different percentiles, where the rows correspond to the environment map (top), a synthetic scene (seen from the top) rendered with (middle) the ground truth environment map and (bottom) our 1-light representation. 
As expected, scenes where the strongest light does not contribute significantly have shadows that are less pronounced which may point to several light sources equally contributing to the overall energy (25th prct.). The strongest light source contributes to more than 80\% of the total energy in at least 50\% of the images in our test set, which confirms our assumption that most scenes can accurately be modeled with a single light source.}
\label{fig:1light-validation}
\end{figure}

\section{Experiments}

\subsection{Validation of our 1-light approximation}

We test our hypothesis that most indoor scenes are well-approximated by a single dominant light source with an ambient term. We render a scene with the ground truth environment map, and compare it with the renders obtained from the parametric lighting optimization procedure described in sec.~\ref{sec:lighting-representation}. Fig.~\ref{subfig:strlight-fig} shows the cumulative distribution of the contribution of the strongest light with respect to the entire lighting of the scene. Note that the strongest light source contributes to more than 95\%/80\%/60\% of the total lighting for 25\%/50\%/75\% of the images in our test set. Fig.~\ref{subfig:strlight-images} shows example images for each of these scenarios. We find that even if we expect indoor scenes to have multiple light sources, the vast majority can accurately be represented by a single \emph{dominant} light. 

\begin{table}[t]
\setlength{\tabcolsep}{4pt}
\centering
\caption{Quantitative comparative metrics on (left) renderings of a diffuse scene, and (right) on the estimated environment maps directly. Each row is color-coded as \colorbox{best}{best} and \colorbox{second}{second best}. We also \colorbox{best}{highlight} the methods which produce lighting representations that can be interpreted and edited by a user (``Edit.''). }

\label{tab:quantitative}
\begin{tabular}{lcccc}
\toprule
& si-RMSE$_\downarrow$
& RMSE$_\downarrow$
& RGB ang.$_\downarrow$
& PSNR$_\uparrow$ \\
\midrule
Ours       
& \cellcolor{best}{0.081} & \cellcolor{best}{0.209} & \cellcolor{second}{4.13$^\circ$} & \cellcolor{best}{12.79} \\ 
Gardner'19 (1)~\cite{Gardner_2019_ICCV} 
& {0.099} & {0.229} & {4.43$^\circ$} & \cellcolor{second}{12.25}  \\ 
Gardner'19 (3)~\cite{Gardner_2019_ICCV} 
& 0.105 & 0.508 & 4.58$^\circ$ & 10.87  \\ 
Gardner'17~\cite{gardner2017learning} 
& 0.123 & 0.628 & 8.29$^\circ$ & 10.24 \\ 
Garon'19~\cite{garon2019fast}
& \cellcolor{second}{0.096} & \cellcolor{second}{0.254} & {8.04$^\circ$} & 9.70  \\ 
Lighthouse~\cite{srinivasan2020lighthouse} 
& 0.120 & 0.253 & 14.53$^\circ$ & 9.88 \\
EMLight~\cite{zhan2021emlight} 
& {0.099} & {0.232} & \cellcolor{best}{3.99$^\circ$} & {10.38} \\
EnvmapNet\footnote{Only their proposed ClusterID loss and tonemapping.}~\cite{somanath2021hdr} 
& {0.097} & {0.286} & {7.67$^\circ$} & {11.74} \\
\bottomrule
\end{tabular}
\begin{tabular}{c}
\toprule
FID$_\downarrow$ \\ 
\midrule
\cellcolor{best}89.58 \\ 
356.8 \\ 
335.6 \\ 
254.8 \\ 
314.9 \\ 
195.5 \\ 
\cellcolor{second}121.09 \\
201.20 \\
\bottomrule
\end{tabular}
\begin{tabular}{c}
\toprule
Edit. \\ 
\midrule
\cellcolor{best}{yes} \\
\cellcolor{best}{yes} \\
\cellcolor{best}{yes} \\
no \\
no \\
no \\
no \\
no \\
\bottomrule
\end{tabular}
\end{table}

%!TEX root = eccv2022submission.tex
\begin{figure}[ht!]
    \scriptsize
    \centering
    \setlength{\tabcolsep}{0.5pt}
    \renewcommand{\arraystretch}{0.5}
    \newlength{\mywidth}
    \setlength{\mywidth}{0.12\linewidth}
    \begin{tabular}{cccccccccc}
    &   
    Input &
    GT &
    % \cite{Gardner_2019_ICCV} (1) & 
    \cite{Gardner_2019_ICCV} (3) &
    \cite{gardner2017learning} &
    \cite{garon2019fast} &
    \cite{srinivasan2020lighthouse} & 
    \cite{zhan2021emlight} & 
    Ours \\
    \rotatebox{90}{\tiny \hspace{1.8em} 1st} &
    \includegraphics[width=\mywidth]{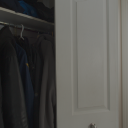} & 
    \includegraphics[width=\mywidth]{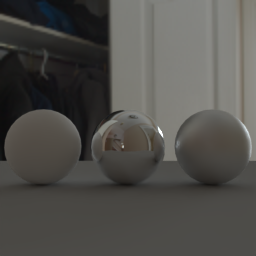} & 
    \includegraphics[width=\mywidth]{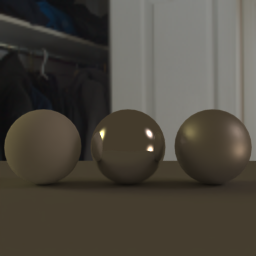} & 
    \includegraphics[width=\mywidth]{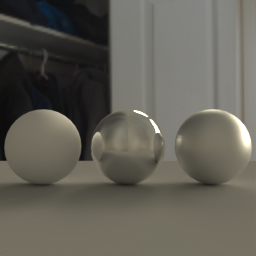} & 
    \includegraphics[width=\mywidth]{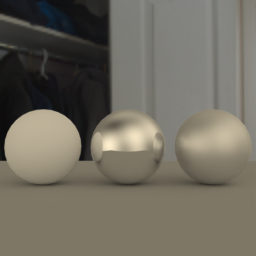} & 
    \includegraphics[width=\mywidth]{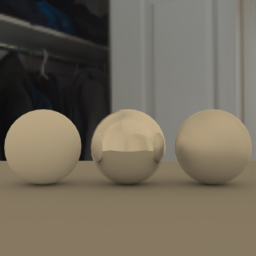} & 
    \includegraphics[width=\mywidth]{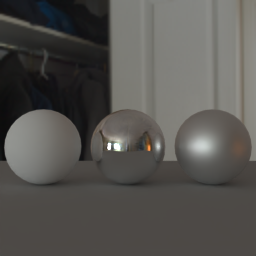} & 
    \includegraphics[width=\mywidth]{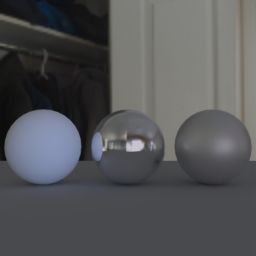} \\ 
    & 
    &
    \includegraphics[width=\mywidth]{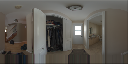} & 
    \includegraphics[width=\mywidth]{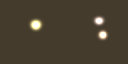} & 
    \includegraphics[width=\mywidth]{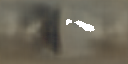} & 
    \includegraphics[width=\mywidth]{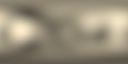} & 
    \includegraphics[width=\mywidth]{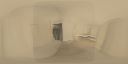} & 
    \includegraphics[width=\mywidth]{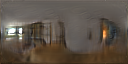} & 
    \includegraphics[width=\mywidth]{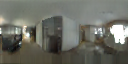} \\ 
    & 
    & 
    & 
    {\tiny 0.03, 7.26$^\circ$} &
    {\tiny 0.52, 2.96$^\circ$} &
    {\tiny 0.02, 1.86$^\circ$} &
    {\tiny 0.24, 1.30$^\circ$} &
    {\tiny 0.03, 2.58$^\circ$} &
    {\tiny 0.03, 6.70$^\circ$} 
    \vspace{3pt} \\
    \rotatebox{90}{\tiny \hspace{1.8em} 25th} &
    \includegraphics[width=\mywidth]{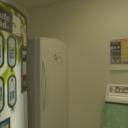} & 
    \includegraphics[width=\mywidth]{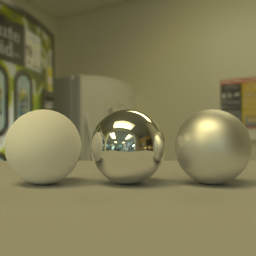} & 
    \includegraphics[width=\mywidth]{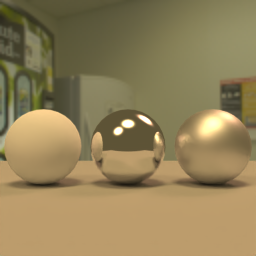} & 
    \includegraphics[width=\mywidth]{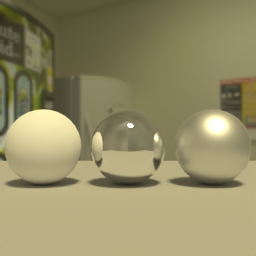} & 
    \includegraphics[width=\mywidth]{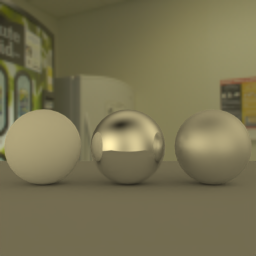} & 
    \includegraphics[width=\mywidth]{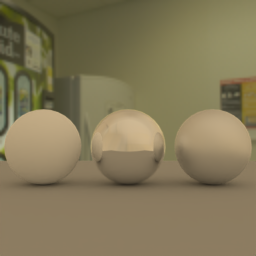} & 
    \includegraphics[width=\mywidth]{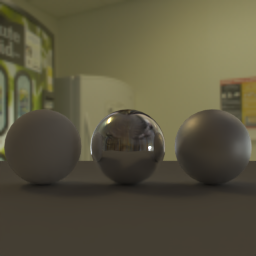} & 
    \includegraphics[width=\mywidth]{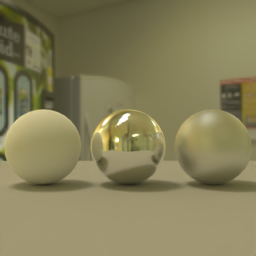} \\ 
    &
    &
    \includegraphics[width=\mywidth]{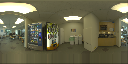} & 
    \includegraphics[width=\mywidth]{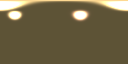} & 
    \includegraphics[width=\mywidth]{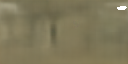} & 
    \includegraphics[width=\mywidth]{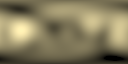} & 
    \includegraphics[width=\mywidth]{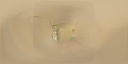} & 
    \includegraphics[width=\mywidth]{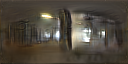} & 
    \includegraphics[width=\mywidth]{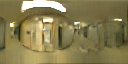} \\ 
    &
    & 
    & 
    {\tiny 0.71, 3.53$^\circ$} &
    {\tiny 1.27, 9.08$^\circ$} &
    {\tiny 0.31, 8.89$^\circ$} &
    {\tiny 0.11, 13.3$^\circ$} &
    {\tiny 0.33, 5.52$^\circ$} &
    {\tiny 0.11, 1.73$^\circ$}
    \vspace{3pt} \\
    \rotatebox{90}{\tiny \hspace{1.8em} 50th} &
    \includegraphics[width=\mywidth]{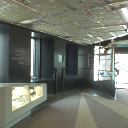} & 
    \includegraphics[width=\mywidth]{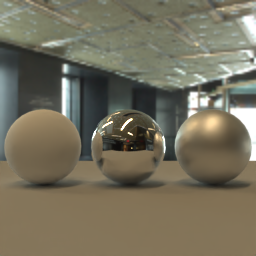} & 
    \includegraphics[width=\mywidth]{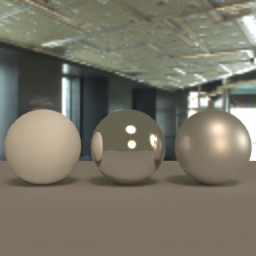} & 
    \includegraphics[width=\mywidth]{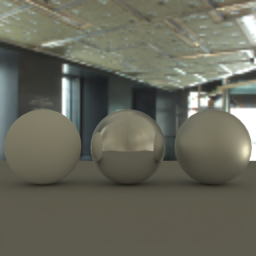} & 
    \includegraphics[width=\mywidth]{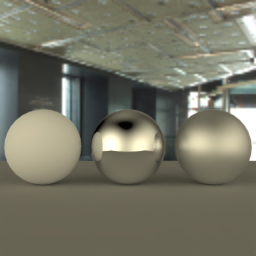} & 
    \includegraphics[width=\mywidth]{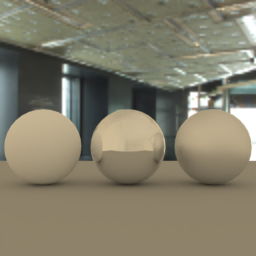} & 
    \includegraphics[width=\mywidth]{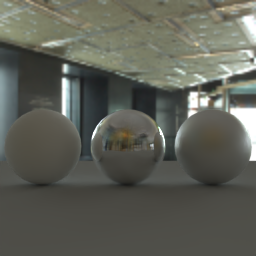} & 
    \includegraphics[width=\mywidth]{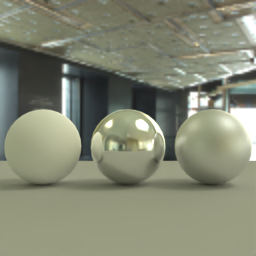} \\ 
    &
    &
    \includegraphics[width=\mywidth]{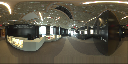} & 
    \includegraphics[width=\mywidth]{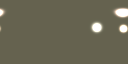} & 
    \includegraphics[width=\mywidth]{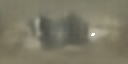} & 
    \includegraphics[width=\mywidth]{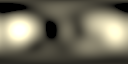} & 
    \includegraphics[width=\mywidth]{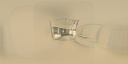} & 
    \includegraphics[width=\mywidth]{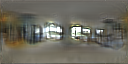} & 
    \includegraphics[width=\mywidth]{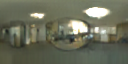} \\ 
    &
    & 
    & 
    {\tiny 0.27, 14.5$^\circ$} &
    {\tiny 0.13, 16.2$^\circ$} &
    {\tiny 0.24, 15.4$^\circ$} &
    {\tiny 0.15, 16.3$^\circ$} &
    {\tiny 0.19, 13.6$^\circ$} &
    {\tiny 0.20, 9.65$^\circ$} 
    \vspace{3pt} \\
    \rotatebox{90}{\tiny \hspace{1.8em} 75th} &
    \includegraphics[width=\mywidth]{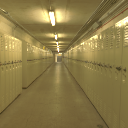} & 
    \includegraphics[width=\mywidth]{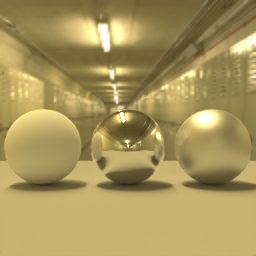} & 
    \includegraphics[width=\mywidth]{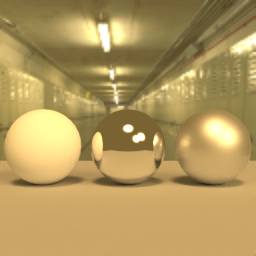} & 
    \includegraphics[width=\mywidth]{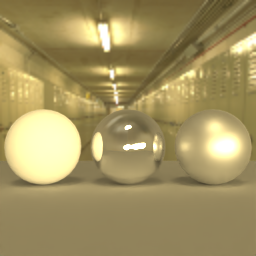} & 
    \includegraphics[width=\mywidth]{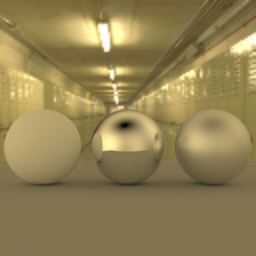} & 
    \includegraphics[width=\mywidth]{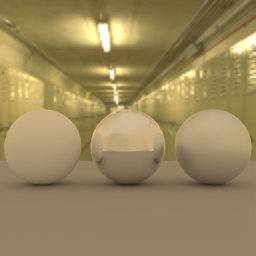} & 
    \includegraphics[width=\mywidth]{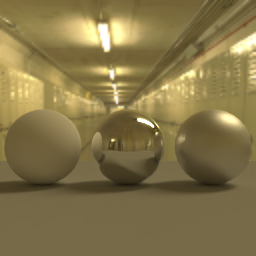} & 
    \includegraphics[width=\mywidth]{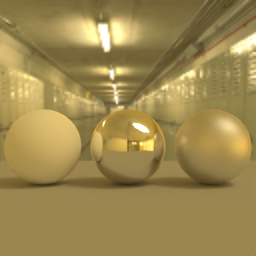} \\ 
    &
    &
    \includegraphics[width=\mywidth]{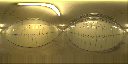} & 
    \includegraphics[width=\mywidth]{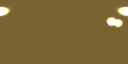} & 
    \includegraphics[width=\mywidth]{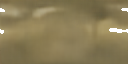} & 
    \includegraphics[width=\mywidth]{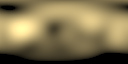} & 
    \includegraphics[width=\mywidth]{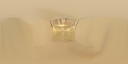} & 
    \includegraphics[width=\mywidth]{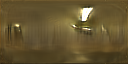} & 
    \includegraphics[width=\mywidth]{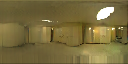} \\ 
    &
    & 
    & 
    {\tiny 0.56, 2.25$^\circ$} &
    {\tiny 0.35, 9.18$^\circ$} &
    {\tiny 0.67, 8.90$^\circ$} &
    {\tiny 0.40, 17.9$^\circ$} &
    {\tiny 0.55, 1.93$^\circ$} &
    {\tiny 0.39, 3.33$^\circ$} \\
    \end{tabular}
    \caption{Qualitative lighting estimation examples from our test set. To compare the estimated lighting, we render a simple scene composed of three spheres (diffuse, mirror, glossy) on a diffuse ground plane with different methods. From left to right: input image, ground truth lighting, Gardner'19~\cite{Gardner_2019_ICCV} (3 lights), Gardner'17~\cite{gardner2017learning}, Garon'19~\cite{garon2019fast}, Lighthouse~\cite{srinivasan2020lighthouse}, EMLight~\cite{zhan2021emlight}, and ours. The second row shows the corresponding estimated lighting in equirectangular format (reprojected in the center of the image for the spatially-varying techniques such as \cite{Gardner_2019_ICCV,garon2019fast,srinivasan2020lighthouse} and ours). Finally, error metrics (RMSE and RGB angular) are also shown below each example for reference. Each group shows examples from different error percentiles for our method according to the RMSE metric. More examples can be found in the supplementary materials.}
    \label{fig:qualitative}
    \end{figure}

\subsection{Light estimation comparison}

We now evaluate our method and compare it with recent state-of-the-art light estimation approaches. We first validate that our model performs better on quantitative metrics evaluated on physic-based renders of a scene using a test set provided by \cite{gardner2017learning}. For each of the 224 panoramas in the test split, we extract 10 images using the same sampling distribution as in \cite{gardner2017learning}, for a total of 2,240 images for evaluation. We also show renders in various scenes to demonstrate how our solution is visually more appealing.

\subsubsection{Quantitative comparison}

To evaluate the lighting estimates, we render a test scene composed of an array of spheres viewed from above (sec.~\ref{sec:lighting-representation}) and compute error metrics on the resulting rendering when compared to the ground truth obtained with the original HDR panorama. We report RMSE, si-RMSE~\cite{grosse2009ground}, PSNR, and RGB angular error~\cite{legendre2019deeplight}. We also compute the FID\footnote{Implementation taken from \url{https://pypi.org/project/pytorch-fid/}.} on the resulting environment maps to evaluate the realism of reflections (similar to \cite{somanath2021hdr}). % Each method is evaluated with the same rendering strategy, (i.e. a render with the estimated envmap), except ours since we have our own representation as described in \ref{sec:lighting-representation}. 

We evaluate against the following works. 
First, two versions of \cite{Gardner_2019_ICCV} are compared: the original (3) where 3 light sources are estimated, and a version (1) trained to predict a single parametric light. Second, we also compare to Lighthouse~\cite{srinivasan2020lighthouse}, which expects a stereo pair as input. As a substitute, we generate a second image with a small baseline using Synsin~\cite{wiles2020synsin} (visual inspection confirmed this yields reasonable results). For \cite{garon2019fast}, we select the coordinates of the image center for the object position. For \cite{somanath2021hdr}, we implemented their proposed ``Cluster ID loss'' and tonemapping (eq.~1 in \cite{somanath2021hdr}) but used pix2pixHD as backbone. Finally, we also compare against \cite{zhan2021emlight}. Results for each metrics are reported in tab.~\ref{tab:quantitative}, which shows that despite our model being simple, it achieves the best score in every metric. We argue that it is \emph{because} of its simplicity that we can achieve competitive results. Our approach can be trained on real data (as opposed to \cite{garon2019fast,srinivasan2020lighthouse} and does not require an elaborate 2-stage training process (compared to \cite{Gardner_2019_ICCV}). 
We also demonstrate a significantly lower FID score than other methods thus bridging the gap between representation realism and HDR accuracy. 

\subsubsection{Qualitative comparison}
\begin{figure}[t!]
\centering
\scriptsize
\setlength{\mywidth}{0.225\linewidth}
\setlength{\tabcolsep}{2pt}
\begin{tabular}{cccc}
\rotatebox{90}{\hspace{.65cm}Input image} &
\includegraphics[width=\mywidth]{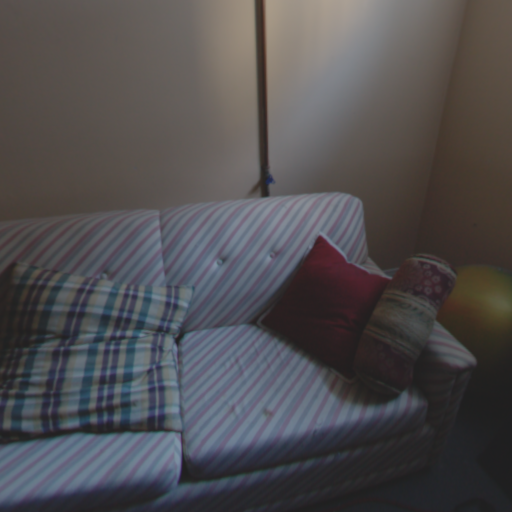} & 
\includegraphics[width=\mywidth]{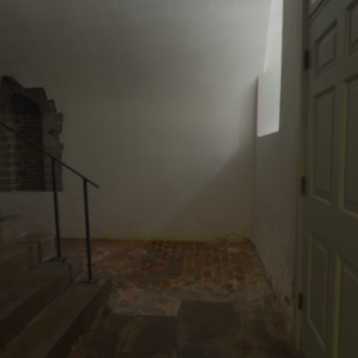} & 
\includegraphics[width=\mywidth]{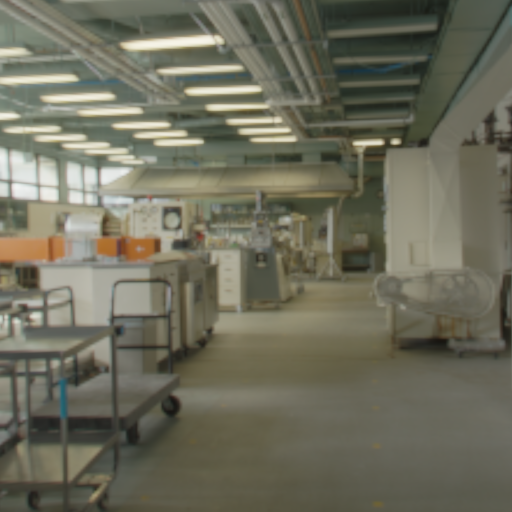} \\
\rotatebox{90}{\hspace{.01cm}Texture $\hat{\mathbf{T}}$} &
\includegraphics[width=\mywidth]{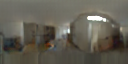} & 
\includegraphics[width=\mywidth]{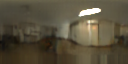} & 
\includegraphics[width=\mywidth]{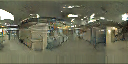} \\ 
\rotatebox{90}{\hspace{.25cm}Rendered objects} &
\includegraphics[width=\mywidth]{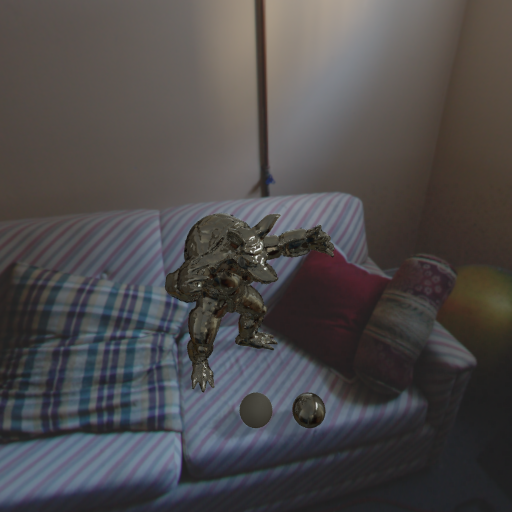} & 
\includegraphics[width=\mywidth]{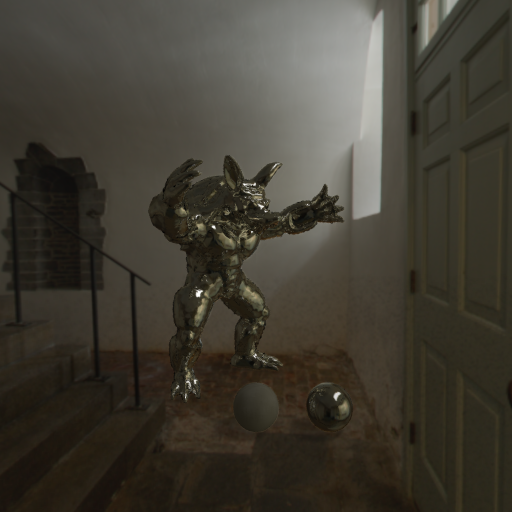} & 
\includegraphics[width=\mywidth]{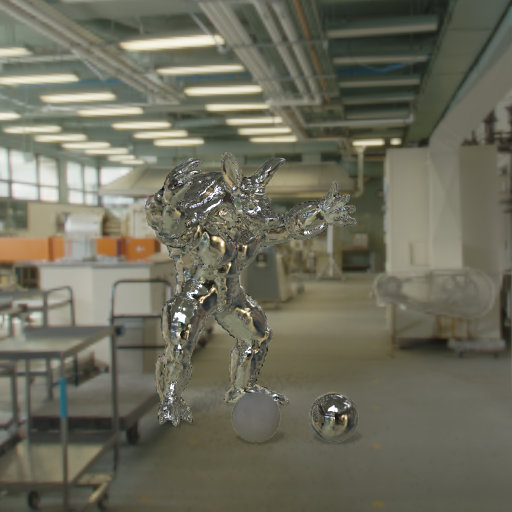} \\
\end{tabular}
\caption{Virtual object insertion in scenes with our estimated lighting. For simplicity, we assume the scene surrounding the objects is made of a flat ground plane, which catches shadows and is placed manually by an artist (the focus of our work being lighting estimation). For example, the figure shows a golden armadillo and sphere inserted into three different scenes. Note how the reflections on the objects and the shadows cast on the ground plans appear realistic.}
\label{fig:blender}
\end{figure}

\begin{figure}[t!]
\centering
\scriptsize
% \renewcommand{\arraystretch}{0.5}
% \newlength{\mywidth}
\setlength{\tabcolsep}{1pt}
\subfloat[Input image]{
	\setlength{\mywidth}{0.2\linewidth}
	\begin{tabular}{cc}
	\includegraphics[width=\mywidth]{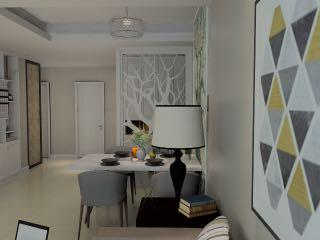} \\ 
	\includegraphics[width=\mywidth]{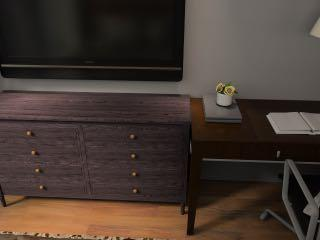} \\
	\end{tabular}
}
\subfloat[Wang et al.]{
	\setlength{\mywidth}{0.2\linewidth}	
	\begin{tabular}{cc}
	\includegraphics[width=\mywidth]{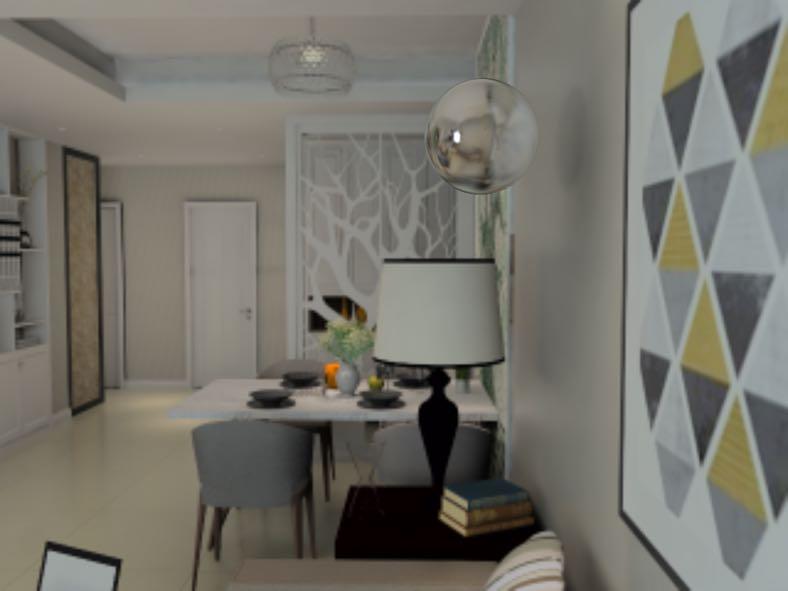} \\
	\includegraphics[width=\mywidth]{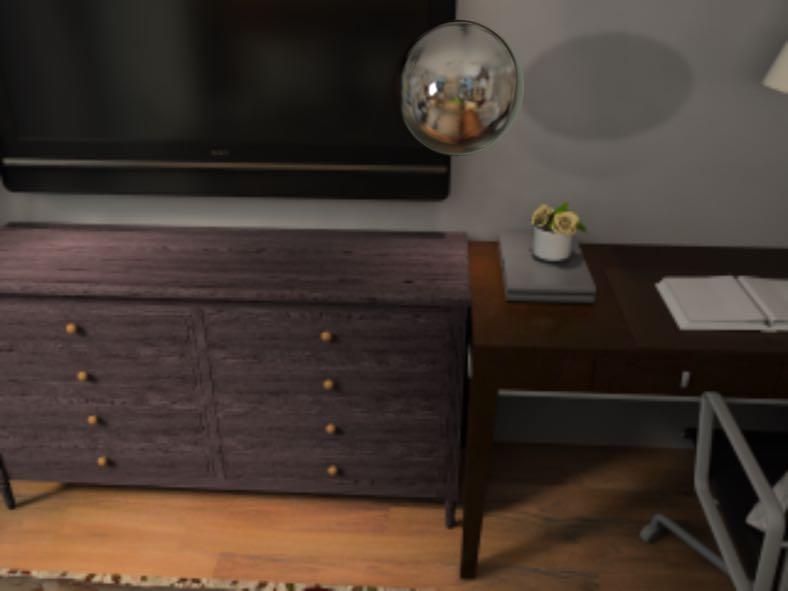} \\
	\end{tabular}
}
\subfloat[Ours]{
	\setlength{\mywidth}{0.2\linewidth}	
	\begin{tabular}{cc}
	\includegraphics[width=\mywidth]{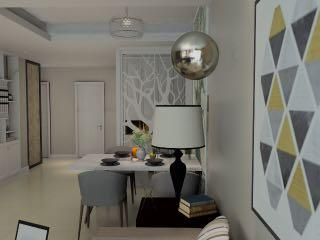} \\
	\includegraphics[width=\mywidth]{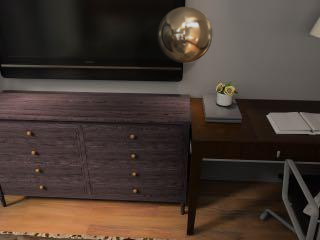} \\
	\end{tabular}
}
\caption[]{Qualitative comparison against Wang et al.~\cite{wang2021learning}}
\label{fig:nvidia}
\end{figure}

We also present qualitative results in fig.~\ref{fig:qualitative} where predictions are rendered on 3 spheres with varying reflectance properties (diffuse, mirror, glossy). In addition, a tonemapped equirectangular view of the estimated light representation is provided under each render. We show an example from each error percentiles according to the RMSE metric. Our proposed method is perceptually better on the mirror spheres as other methods do not model high frequency details from the scene. We also notice accurate shadow and shading from all the spheres. We show objects realistically composited into photographs in fig.~\ref{fig:blender}. Note how the reflections of the virtual objects and their cast shadows on the ground plane perceptually match the input photograph. Finally, we also compare against~\cite{wang2021learning} in fig.~\ref{fig:nvidia}.

%!TEX root = eccv2022submission.tex

%!TEX root = eccv2022submission.tex

%!TEX root = eccv2022submission.tex
\begin{figure}[t]
\centering
\tiny
\setlength{\mywidth}{0.22\linewidth}	
\begin{tabular}{ccccc}
& Layout from \cite{yang2022learning} & No layout & Layout from \cite{yang2022learning} & No layout \\
\rotatebox{90}{\hspace{2.5em} Image}& 
\includegraphics[width=\mywidth]{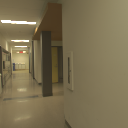} &
\includegraphics[width=\mywidth]{imgs/black_layout/9C4A0566-a088c98ccf_00/crop/9C4A0566-a088c98ccf_00.png} &
\includegraphics[width=\mywidth]{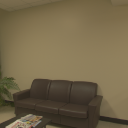} &
\includegraphics[width=\mywidth]{imgs/black_layout/9C4A1774-12d053d172_00/crop/9C4A1774-12d053d172_00.png} \\
\rotatebox{90}{\hspace{2em} Image layout}& 
\includegraphics[width=\mywidth]{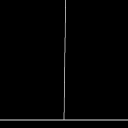} &
\includegraphics[width=\mywidth]{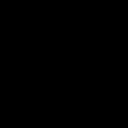} &
\includegraphics[width=\mywidth]{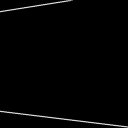} &
\includegraphics[width=\mywidth]{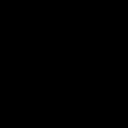} \\
\rotatebox{90}{\hspace{0.2em} Texture $\hat{\mathbf{T}}$} &
\includegraphics[width=\mywidth]{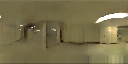} & 
\includegraphics[width=\mywidth]{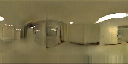} &
\includegraphics[width=\mywidth]{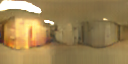} & 
\includegraphics[width=\mywidth]{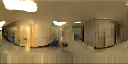} \\
\rotatebox{90}{\hspace{0.3em} Layout $\hat{\mathbf{L}}$} &
\includegraphics[width=\mywidth]{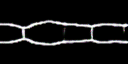} & 
\includegraphics[width=\mywidth]{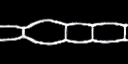} &
\includegraphics[width=\mywidth]{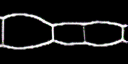} & 
\includegraphics[width=\mywidth]{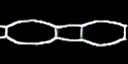} \\
\end{tabular}
\caption{Ablation on input image layout. We compare the output of our method (last two rows) as a function of whether or not it is given the estimated layout of the input image (with \cite{yang2022learning}). Our approach produces similar results in both cases. }
\label{fig:nolayout}
\end{figure}

\subsection{Ablation study on input layout}

One may consider that requiring the image layout as input may make our method sensitive to its estimation. To show this is not the case, we perform an experiment where we provide a black layout as input to the layout network (equivalent to no layout estimate). As can be seen in fig.~\ref{fig:nolayout}, providing a black layout as input simply results in a different layout prediction where the texture still remains coherent with the RGB input and estimated light direction. The FID of the generated panoramas with no input layout is 88.68 (compared to 89.58 from tab.~\ref{tab:quantitative}), showing that this essentially has no impact. 

\subsection{Ablation study on the texture network}

We also tested different configurations for the texture network in order to validate our design choices. More specifically, we trained the texture network providing as input: (1) only the RGB crop (FID of 167.39), (2) RGB crop and parametric light (FID of 97.13), and (3) RGB crop and layout (FID of 151.04). In contrast, our full approach obtained an FID of 89.57 (see tab.~\ref{tab:quantitative}).
%!TEX root = eccv2022submission.tex
\section{Editing the estimated lighting}

%!TEX root = eccv2022submission.tex
\begin{figure}[t!]
\centering
\scriptsize
% \renewcommand{\arraystretch}{0.5}
% \newlength{\mywidth}
\setlength{\tabcolsep}{1pt}
\subfloat[Light azimuth\label{subfig:edit-azimuth}]{
	\setlength{\mywidth}{0.18\linewidth}
	\begin{tabular}{ccccc}
	\includegraphics[width=\mywidth]{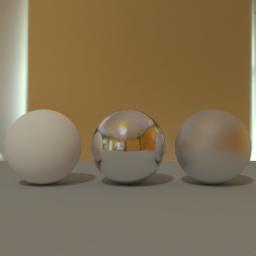} & 
	\includegraphics[width=\mywidth]{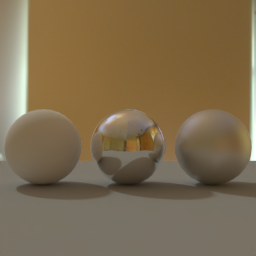} & 
	\includegraphics[width=\mywidth]{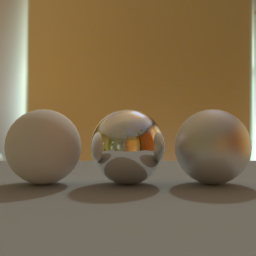} & 
	\includegraphics[width=\mywidth]{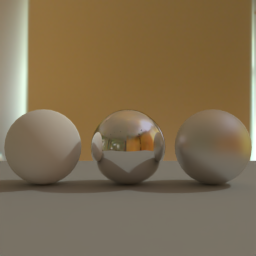} &
	\includegraphics[width=\mywidth]{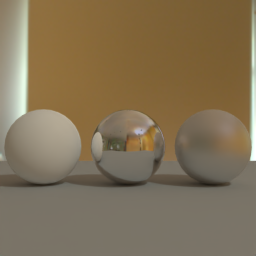} \\ 
	\includegraphics[width=\mywidth]{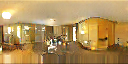} & 
	\includegraphics[width=\mywidth]{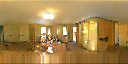} & 
	\includegraphics[width=\mywidth]{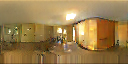} & 
	\includegraphics[width=\mywidth]{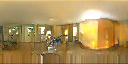} & 
	\includegraphics[width=\mywidth]{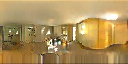} \\ 
	\includegraphics[width=\mywidth]{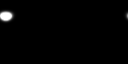} & 
	\includegraphics[width=\mywidth]{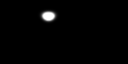} & 
	\includegraphics[width=\mywidth]{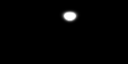} & 
	\includegraphics[width=\mywidth]{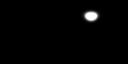} &
	\includegraphics[width=\mywidth]{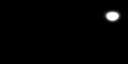} \\
	\end{tabular}
} 
\\
\subfloat[Light elevation\label{subfig:edit-elevation}]{
	\setlength{\mywidth}{0.142\linewidth}
	\begin{tabular}{cc}
	\includegraphics[width=\mywidth]{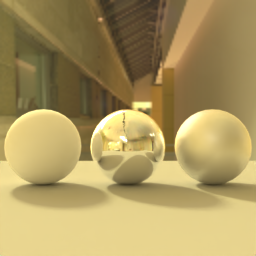} & 
	\includegraphics[width=\mywidth]{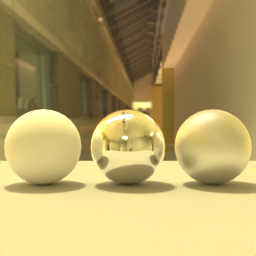} \\ 
	\includegraphics[width=\mywidth]{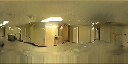} & 
	\includegraphics[width=\mywidth]{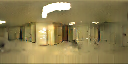} \\ 
	\includegraphics[width=\mywidth]{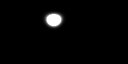} & 
	\includegraphics[width=\mywidth]{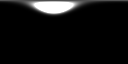} \\ 
	\end{tabular}
}
\subfloat[Light size\label{subfig:edit-size}]{
	\setlength{\mywidth}{0.142\linewidth}	
	\begin{tabular}{cc}
	\includegraphics[width=\mywidth]{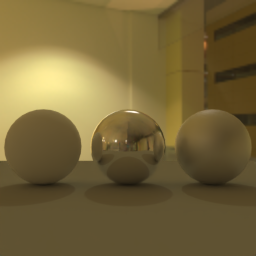} & 
	\includegraphics[width=\mywidth]{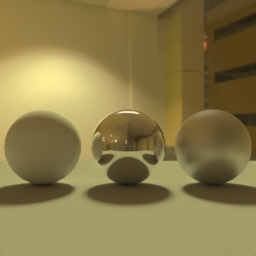} \\
	\includegraphics[width=\mywidth]{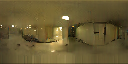} & 
	\includegraphics[width=\mywidth]{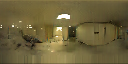} \\
	\includegraphics[width=\mywidth]{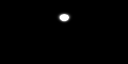} & 
	\includegraphics[width=\mywidth]{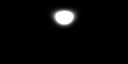} \\
	\end{tabular}
}
\subfloat[Scene layout\label{subfig:edit-layout}]{
	\setlength{\mywidth}{0.142\linewidth}	
	\begin{tabular}{cc}
	\includegraphics[width=\mywidth]{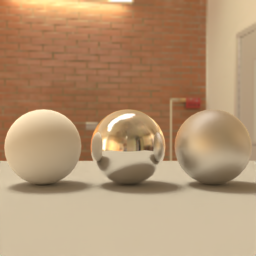} &
	\includegraphics[width=\mywidth]{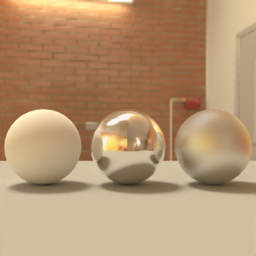} \\
	\includegraphics[width=\mywidth]{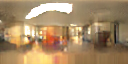} &
	\includegraphics[width=\mywidth]{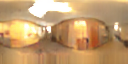} \\
	\includegraphics[width=\mywidth]{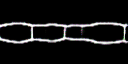} &
	\includegraphics[width=\mywidth]{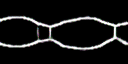} \\
	\end{tabular}
}
\caption[]{Using our representation, a user can easily edit the estimated light parameters and obtain relighting results consistent with their edits. For example, the user can change the \subref{subfig:edit-azimuth} azimuth and \subref{subfig:edit-elevation} elevation angles of the light source; \subref{subfig:edit-size} the size of the light source; or \subref{subfig:edit-layout} the layout of the scene. For all scenarios, we show rendered virtual objects in the first row, the estimated texture $\hat{\mathbf{T}}$ in the second, and the representation being edited in the last (light parameters $\hat{\mathbf{p}}$ for \subref{subfig:edit-azimuth}--\subref{subfig:edit-size} and layout $\hat{\mathbf{L}}$ for \subref{subfig:edit-layout}). }
\label{fig:editability}
\end{figure}

Because of its intuitive nature, it is simple and natural for a user to edit our estimated lighting representation, should the estimate not perfectly match the background image or simply for artistic purposes. Fig.~\ref{fig:editability} shows that our approach simultaneously  \textit{disentangles} various components of illumination, allows an \textit{intuitive control} over those components, and enables \textit{realistic relighting results}. First, fig.~\ref{subfig:edit-azimuth} shows that a user can rotate the light source about its azimuth angle. Note how the estimated texture (second row) is consistent with the desired light position (third row), while preserving the same overall structure. The renders (first row) exhibit realistic reflections and shadows that correspond to the desired lighting directions. A similar behaviour can be observed in figs~\ref{subfig:edit-elevation} and \ref{subfig:edit-size} when the elevation angle and size are modified, respectively. In fig.~\ref{subfig:edit-layout}, we show that it is also possible to edit the scene layout and obtain an estimated texture map $\hat{\mathbf{T}}$ that is consistent with the users request. We also show results of compositing virtual objects directly into a scene in fig.~\ref{fig:blender}. As shown in fig.~\ref{fig:teaser}, realistic rendering results can intuitively be edited to achieve the desired look. 
To the best of our knowledge, the only other method which allows intuitive editing of indoor lighting estimate is that of Gardner et al.~\cite{Gardner_2019_ICCV}. Unfortunately, realistic renders are limited to diffuse objects and cannot be extended to reflective objects as shown in fig.~\ref{fig:qualitative}.

\section{Discussion}

This paper proposes a lighting estimation approach which produces an intuitive, user-editable lighting representation given a single indoor input image. By explicitly representing the dominant light source using a parametric model, and the ambient environment map using a textured cuboid, our approach bridges the gap between generating realistic shading (produced by HDR light sources) and reflections (produced by textured environment maps) on rendered virtual objects. We demonstrate, through extensive experiments, that our approach provides competitive quantitative performance when compared to recent lighting estimation techniques. In particular, when compared to the only other approach which can be user-edited~\cite{Gardner_2019_ICCV}, our approach yields significant improved results. 

\paragraph{Limitations and future work} While our proposed approach estimates a 3D representation of the surrounding lighting environment, it does not reason about light occlusions in the scene as opposed to other techniques such as \cite{garon2019fast,li2020inverse,wang2021learning}. Incorporating these higher-order interactions while maintaining interpretability and editability of the output representation is an interesting direction for future research. In addition, the estimated environment textures were shown to produce realistic reflections on shiny objects, but a close inspection reveals that they are low resolution and contain some visual artifacts. It is likely that more recent image-to-image translation architectures~\cite{choi2020stargan,park2020contrastive} could be used to improve realism.

\paragraph{Acknowledgements} This research was supported by MITACS and the NSERC grant RGPIN-2020-04799. The authors thank Pascal Audet for his help.

% \clearpage
\bibliographystyle{splncs04}
\bibliography{eccv2022cameraready}
\end{document}